\documentclass[letterpaper, 10 pt, conference]{ieeeconf}  
\IEEEoverridecommandlockouts                              

\usepackage{graphics} 
\usepackage{epsfig} 
\usepackage{mathptmx} 
\usepackage{times} 
\usepackage{amsmath} 
\usepackage{amssymb}  
\usepackage{url}

\newcommand{\Brac}[1]{\left\{ #1 \right\}}
\newcommand{\brac}[1]{\left[ #1 \right]}
\newcommand{\paren}[1]{\left( #1 \right)}

\title{\LARGE \bf
Optimal Energy Shaping Control for a Backdrivable Hip Exoskeleton
}

\author{Jiefu Zhang$^{1}$, Jianping Lin$^{1,2}$, Vamsi Peddinti$^{3}$, and Robert D. Gregg$^{3}$
\thanks{*This work was supported by the National Science Foundation under Award Number 1949869 and by the National Institute of Biomedical Imaging and Bioengineering of the NIH under Award Number R01EB031166. The content is solely the responsibility of the authors and does not necessarily represent the official views of the NSF or NIH.}
\thanks{$^{1}$Electrical Engineering and Computer Science,
        University of Michigan, Ann Arbor, MI 48109, USA. $^{2}$State Key Laboratory of Mechanical System and Vibration, School of Mechanical Engineering, Shanghai Jiao Tong University, Shanghai 200240, China. $^{3}$Robotics, University of Michigan, Ann Arbor, MI 48109, USA.
        }
\thanks{Corresponding author: Robert D. Gregg. Contact: {\tt\footnotesize \{zjiefu, jplin, rdgregg\}@umich.edu}}%
}

\begin{document}

\maketitle
\thispagestyle{empty}
\pagestyle{empty}

\begin{abstract}

Task-dependent controllers widely used in exoskeletons track predefined trajectories, which overly constrain the volitional motion of individuals with remnant voluntary mobility. Energy shaping, on the other hand, provides task-invariant assistance by altering the human body's dynamic characteristics in the closed loop. While human-exoskeleton systems are often modeled using Euler-Lagrange equations, in our previous work we modeled the system as a port-controlled-Hamiltonian system, and a task-invariant controller was designed for a knee-ankle exoskeleton using interconnection-damping assignment passivity-based control. In this paper, we extend this framework to design a controller for a backdrivable hip exoskeleton to assist multiple tasks. A set of basis functions that contains information of kinematics is selected and corresponding coefficients are optimized, which allows the controller to provide torque that fits normative human torque for different activities of daily life. Human-subject experiments with two able-bodied subjects demonstrated the controller's capability to reduce muscle effort across different tasks.
\end{abstract}

\section{Introduction}

Lower-limb exoskeletons have proved to be powerful in rehabilitation and restoring mobility, while their controller design remains a challenge. Most commercial exoskeletons like ReWalk and Ekso Bionics \cite{baud2021review} fall into task-dependent controllers tracking predefined trajectories, which are not appropriate for people with remnant voluntary mobility. Besides, the need of detecting users’ intention for the transition between task-dependent controllers makes it difficult to perform a continuum of tasks and may cause injury when detection goes wrong. Moreover, the controller parameter tuning is a laborious, technical challenge, which hinders applying exoskeletons to a larger population.

To overcome these limitations, task-independent control frameworks have been introduced. In \cite{nagarajan2016integral}, an integral admittance shaping controller for single degree-of-freedom (DoF) exoskeletons was proposed, which provided assistance by modifying the dynamic response of the coupled system. Experiments showed that larger motion can be achieved with the same muscle effort. Based on delayed output feedback control, a unified controller was designed in \cite{lim2019delayed}, which is capable to provide assistance under various walking speeds and environments. Learning-based methods have also been investigated in \cite{molinaro2022subject}, which allowed subject-independent hip joint moment estimation over different tasks based on wearable sensor data. However, the ``black box" nature of these learning-based algorithms make it difficult to guarantee safety outside the training dataset.

As a trajectory-free control method, energy shaping provides task-invariant assistance by altering the human body's dynamics in the closed-loop system and has been extensively investigated for exoskeleton control \cite{lv2018design}. In \cite{lv2017underactuated}, a potential energy shaping-based control method was proposed to provide body-weight support (BWS) for exoskeletons using a controlled Lagrangian. However, the control law depends on contact conditions, which change with different gait phases. A unified controller was proposed in \cite{lin2019contact}, which provides task-invariant assistance with respect to different human input and contact conditions. 
While these potential energy shaping methods only provided BWS, total energy shaping can further regulate velocity by modifying the mass/inertia matrix, which was investigated in \cite{lv2021trajectory}. The ability to provide greater assistance compared with potential energy shaping alone was shown by simulation. However, the control law requires computationally-intensive inversion of the shaped mass/inertia matrix, which is also susceptible to singularities due to underactuation.

While the above energy shaping strategies used the controlled Lagrangian method, one can also model the human-exoskeleton system as a port-controlled Hamiltonian system. Then the interconnection and damping assignment passivity-based control (IDA-PBC) method can provide extra shaping freedom compared with the controlled Lagrangian counterpart \cite{ortega2002interconnection}. By altering the interconnection structure of the port-controlled Hamiltonian equations, additional velocity-dependent assistance can be provided without modifying the mass/inertia matrix.
The IDA-PBC method has been applied to control a knee-ankle exoskeleton in \cite{lin2020optimal} and \cite{lin2022optimally}, which proved its ability to achieve task-invariant control for primary activities of daily life (ADL). In this paper, we extend the method in \cite{lin2022optimally} and design a task-invariant controller for a commercial hip exoskeleton using IDA-PBC. Hip exoskeletons do not have access to kinematic information of knee and ankle joints, which makes the previous kinematic models no longer suitable. Instead of modeling the legs separately as in \cite{lin2020optimal} and \cite{lin2022optimally}, we adopt a complete point-footed biped model including a trunk, stance leg, and swing leg. Since the controller is based on the same model for both stance and swing leg, this obviates the need for a foot force sensor to switch between control laws for different legs.

The contributions of this paper are summarized as follows. First, the IDA-PBC method is extended to control a hip exoskeleton using the unified system model for both stance and swing phase. Second, the proposed controller only needs data from onboard sensors (hip joint encoders and thigh IMU), which makes it more practical in daily life settings. 
We begin in Section \ref{sec2} by modeling the human-exoskeleton system and reviewing the corresponding matching condition. In Section \ref{sec3}, a passivity-based data-driven method is used to optimize the controller to fit normative human torques. The hardware implementation and human subject experiment are presented in Section \ref{sec4}, showing that the controller is capable of assisting multiple tasks. Finally, Section \ref{sec5} concludes the paper. 



\section{Energy Shaping of Human-Exoskeleton System}
\label{sec2}
In this section, we model the human-exoskeleton system as a port-controlled Hamiltonian system and give the energy shaping-based control law. We also review the matching condition for feasible control laws derived in \cite{lin2022optimally}.

\begin{figure}[htbp]
\centering
\begin{tabular}{ll}
\includegraphics[width = 0.18\textwidth]{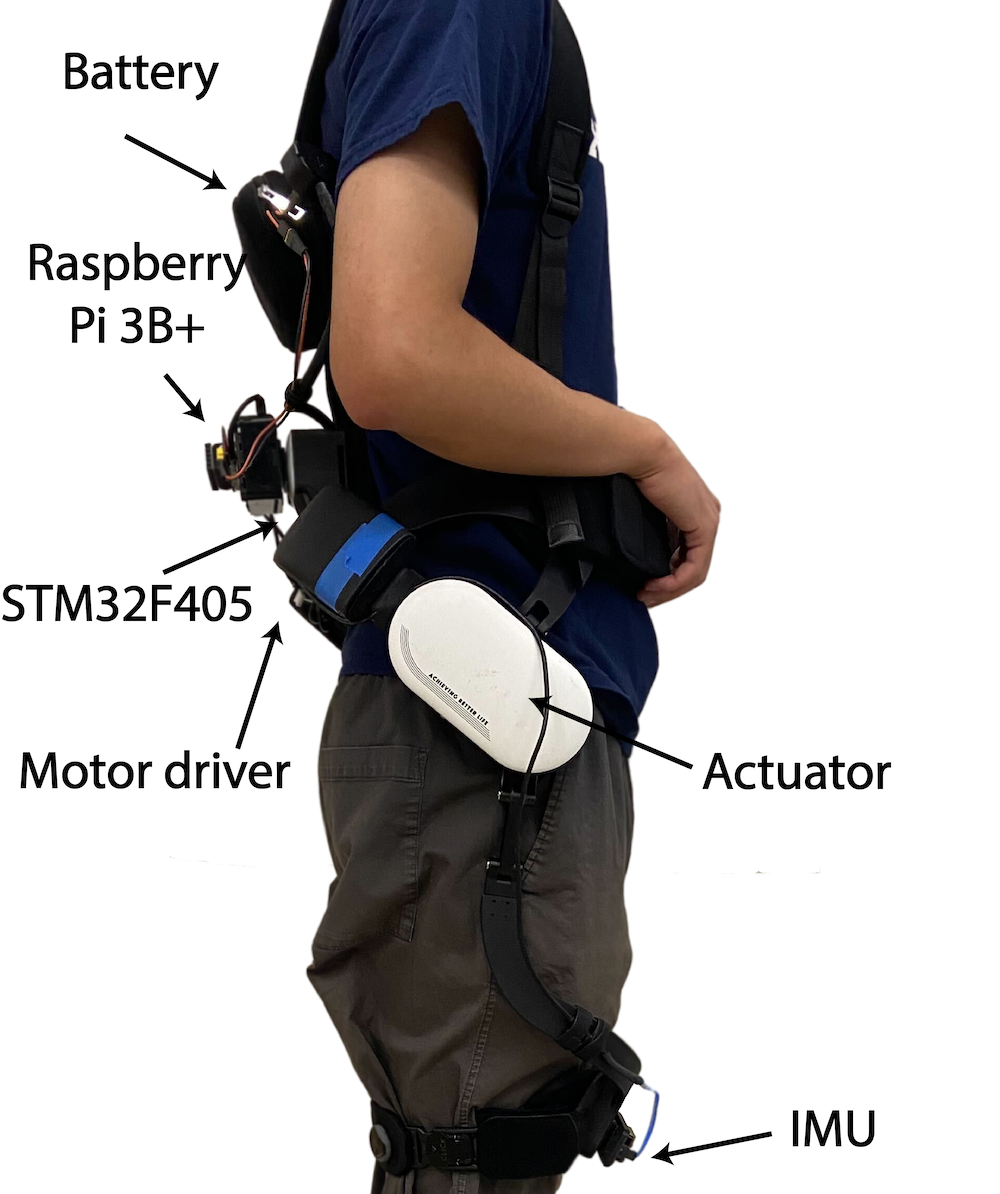}
&
\includegraphics[width = 0.24\textwidth]{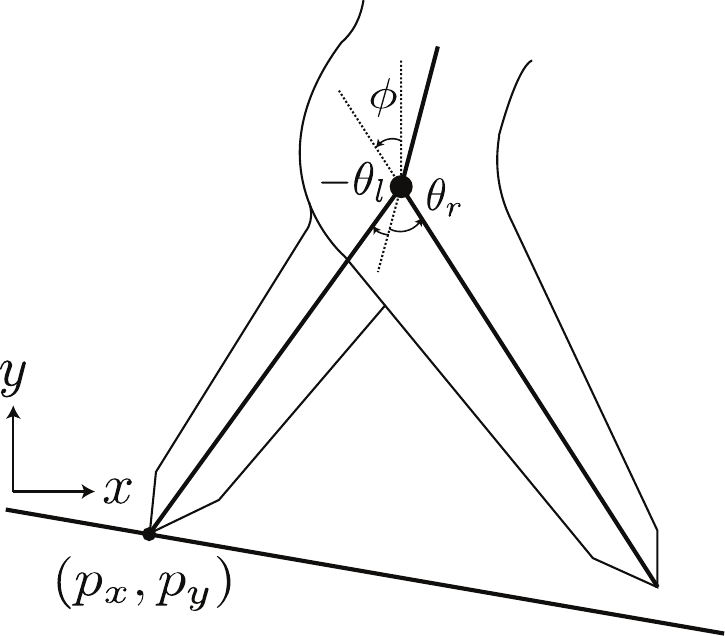}
\end{tabular}
\caption{Left: \emph{Movex} hip exoskeleton produced by Enhanced Robotics. The Raspberry Pi and IMU are added for research proposes. Right: Kinematic model of human-exoskeleton system.}
\label{Fig:model}
\end{figure}

\subsection{System Modeling}

Consider the 5-degree of freedom (DoF) human-exoskeleton system shown in Fig. \ref{Fig:model}. The Cartesian coordinate of stance feet $(p_x, p_y)$ is defined with respect to the inertial reference frame (IRF). The angle between left thigh and trunk is defined as $\theta_l$, and the angle between right thigh and trunk is defined as $\theta_r$. The global thigh angle $\phi$ is defined as the angle between right thigh and the vertical axis. The generalized coordinate of the model is $q = [p_x, p_y, \phi, \theta_l, \theta_r] \in \mathbb{R}^{5}$ in the configuration space $\mathcal{Q} = \mathbb{R}^{5}.$ Define the conjugate momenta as $p = M(q)\dot{q} \in \mathbb{R}^{5}$, where $M(q) \in \mathbb{R}^{5 \times 5}$ is the positive definite mass/inertia matrix and $\dot{q} \in \mathbb{R}^{5}$ is the generalized velocity vector. Then we obtain the port-controlled Hamiltonian (PCH) system characterized by the Hamiltonian: $H(q, p) = \frac{1}{2}p\,M^{-1}(q) p + V(q)$, where $V(q)$ is the potential energy. The state-space form of the PCH dynamics can be given as
\begin{equation}
    \begin{bmatrix}
    \dot{q} \\ \dot{p}
    \end{bmatrix} = \begin{bmatrix}
    0_{5 \times 5} & I_{5 \times 5} \\ -I_{5 \times 5} & 0_{5 \times 5}
    \end{bmatrix} \nabla H + \begin{bmatrix}
    0 \\ \tau + A^T\lambda
    \end{bmatrix},
\label{HamModel}
\end{equation}
where $\nabla H = [\nabla_q H, \nabla_p H] \in \mathbb{R}^{10}$ is the gradient of the Hamiltonian. The torque $\tau = \tau_{exo} + \tau_{hum} \in \mathbb{R}^{5}$ is the sum of exoskeleton input $\tau_{exo} = B u$ and human input $\tau_{hum} = B v$, where $u, v \in \mathbb{R}^{2}$ are the exoskeleton and human input torque applied to hip joints and $B = [0_{2 \times 3}, I_{2 \times 2}]^T\in \mathbb{R}^{5 \times 2}$ is the mapping matrix. Since the number of actuated coordinates is less than the number of generalized coordinates, the system is underactuated. For sake of simplicity, we omit $q$ and $p$ from now on.

The holonomic contact constraints in the system \eqref{HamModel} can be expressed as $a(q) = 0_{c \times 1}$, where $c$ is the number of constraints. This can be written in a matrix form $a(q) = A(q)\,q $. Since $a(q) = 0$ is independent of time, $A(q)$ can be obtained by solving $\dot{a}(q) = \nabla a(q) \dot{q} = A(q)\, \dot{q} = 0$. In our case, $a(q) = [p_x, p_y]^T = 0_{1 \times 2}$, $A = [A_c, 0_{2 \times 2}] = [I_{2 \times 2}, 0_{2 \times 3}]$. The Lagrangian multiplier $\lambda \in \mathbb{R}^{2}$ represents the ground reaction forces, which is mapped to the system through $A$. By differentiating $A\,\dot{q} = 0$ along time and plug \eqref{HamModel} into the equation, we obtain $\lambda$ as
\begin{equation*}
\begin{aligned}
\lambda = & W  \Big\{-\nabla_q \brac{A \paren{\nabla_p H}^T} \paren{\nabla_p H}^T + A \paren{\nabla^2_{p^2}H} \\
& \times \brac{\paren{\nabla_q H}^T - \tau} \Big\} \ ,
\end{aligned}
\end{equation*}
where $W = \brac{A \paren{\nabla^2_{p^2} H} A^T}^{-1} \in \mathbb{R}^{2 \times 2}$. 

\subsection{Review of Matching Conditions for Port-Controlled Hamiltonian System}

In this part, we briefly review the matching condition of a 5-DoF system derived in \cite{lin2022optimally}. Consider the closed-loop system with $u$ being controlled while $v$ remains open-looped. The desired closed-loop Hamiltonian $\tilde{H} = \frac{1}{2} p^T \tilde{M}^{-1} p + \tilde{V}$, where $\tilde{V} = V + \hat{V}$ represents the desired potential energy with shaping term $\hat{V}$. Therefore, $\tilde{N} = \paren{\nabla_q \tilde{H}}^T = \paren{\nabla_q H}^T + \paren{\nabla_q \hat{H}}^T = N + \hat{N}$, where $N$ and $\hat{N}$ represents the corresponding gravitational vectors. We let $\tilde{M} = M$ remain unchanged, which simplifies the matching process. However, with certain structure of the interconnection matrix being satisfied, velocity-dependent shaping can still be achieved \cite{ortega2001putting}. 

Consider the desired closed-loop system
\begin{equation}
\label{DesHam}
    \begin{bmatrix}
    \dot{q} \\ \dot{p}
    \end{bmatrix} = \begin{bmatrix}
    0 & I \\ -I & J_2
    \end{bmatrix} \nabla \tilde{H} + \begin{bmatrix}
    0 \\ B v + A^T \tilde{\lambda} + T_{ext}
    \end{bmatrix},
\end{equation}
where $T_{ext}$ denotes the external input which helps preserve the passivity of \eqref{DesHam}. $J_2 = -J_2^T \in \mathbb{R}^{5 \times 5}$ is skew-symmetric defined as $J_2 = \brac{\paren{\nabla_q p}^T - \paren{\nabla_q p}}+ \brac{\paren{\nabla_q Q}^T - \nabla_q Q} =  \paren{\nabla_q Q}^T - \nabla_q Q$
, since $\nabla_q\,p = 0$. $Q(q) \in \mathbb{R}^5$ is any smooth vector-valued function with respect to $q$ which allows it to provide extra shaping DoF \cite{blankenstein2002matching}. The closed-loop GRF can then be expressed as
\begin{equation*}
    \begin{aligned}
    \tilde{\lambda} = & W  \Big\{-\nabla_q \brac{A \paren{\nabla_p H}^T} \paren{\nabla_p H}^T + A \paren{\nabla^2_{p^2}H}  \\
    & \cdot \brac{\paren{\nabla_q \tilde{H}}^T - J_2 \paren{\nabla_p H}^T - B v - T_{ext}} \Big\} \ .
    \end{aligned}
\end{equation*}

By the standard form of matching condition in \cite{ortega2013passivity}, system \eqref{HamModel} and \eqref{DesHam} match if 
\begin{equation}
\label{matching1}
\begin{aligned}
-&\nabla_q H + B\paren{u+v} + A^T \lambda \\
    &= -\nabla_q \tilde{H} + J_2 \nabla_p \tilde{H} + B v + A^T \tilde{\lambda} + T_{ext}\ .
\end{aligned}
\end{equation}

Plugging $\lambda$ and $\tilde{\lambda}$ into \eqref{matching1} gives
\begin{equation*}
    \begin{aligned}
        B u = &\paren{\nabla_q H}^T - \paren{\nabla_q \tilde{H}}^T + J_2 \nabla_p \tilde{H} + T_{ext} \\
        &+ A^T \Big\{W A \paren{\nabla_{p^2}^2 H} \big[\paren{-\nabla_q H}^T + \paren{\nabla_q \tilde{H}}^T \\
        &- J_2 \nabla_p \tilde{H} + T_{ext}\big]\Big\} \ ,
    \end{aligned}
\end{equation*}
which can be rewritten as
\begin{equation*}
    B_{\lambda} u = X_{\lambda}\brac{\paren{\nabla_q H}^T - \paren{\nabla_q \tilde{H}}^T + J_2 M^{-1}p + T_{ext}}\ ,
\end{equation*}
where $X_{\lambda} = I - A^T W A \paren{\nabla_{p^2}^2 H}$ and $B_{\lambda} = X_{\lambda}B$. Note that here we used $\paren{\nabla_p \tilde{H}}^T = \paren{\nabla_p H}^T = M^{-1} p$ to simplify the equation. 

The matching condition of \eqref{HamModel} and \eqref{DesHam} can be given as
\begin{equation}
\label{matching2}
    0 = B^{\perp}_{\lambda} X_{\lambda}\brac{\paren{\nabla_q H}^T - \paren{\nabla_q \tilde{H}}^T + J_2 M^{-1} p + T_{ext}}\ ,
\end{equation}
where $B^{\perp}_{\lambda} \in \mathbb{R}^{3 \times 5}$ is the full-rank left annihilator of $B_{\lambda}$ such that $B_{\lambda}^{\perp} B_{\lambda} = 0 \ $\cite{blankenstein2002matching}. To solve the matching condition \eqref{matching2}, we first decompose $M$ into submatrices as in \cite{lin2020optimal}:
\begin{equation*}
    M = \begin{bmatrix}
    M_1 & M_2 \\
    M_2^T & M_4
    \end{bmatrix}\ ,
\end{equation*}
where $M_1 \in \mathbb{R}^{3 \times 3}$ corresponds to the unactuated parts $\paren{p_x, p_y, \phi}$ and $M_4 \in \mathbb{R}^{2 \times 2}$ corresponds to the actuated joints $\paren{\theta_l, \theta_r}$. The Schur complement of $M_4$ is $\Delta = M_1 - M_2 M_4^{-1} M_2^T$ and we have $\det \paren{M} = \det \paren{M_4} \det \paren{\Delta}$, which implies that $\Delta$ is nonsingular since $M$ and $M_4$ are nonsingular. Then we obtain
\begin{equation*}
    M^{-1} = \begin{bmatrix}
    \Delta^{-1} & -\Delta^{-1} M_2 M_4^{-1} \\
    -M_4^{-1} M_2^T \Delta^{-1} & M_4^{-1} + M_4^{-1} M_2^T \Delta^{-1} M_2 M_4^{-1}
    \end{bmatrix}.
\end{equation*}

Therefore, we have $W = A_c \Delta^{-1} A_c^T$ and $X_{\lambda}$ can be rewritten as
\begin{equation}
\label{X_lambda}
    X_{\lambda} = \begin{bmatrix}
    I_{3 \times 3} - Z_{\lambda} & Z_{\lambda}M_2 M_4^{-1} \\
    0_{2 \times 3} & I_{2 \times 2}
    \end{bmatrix}\ ,
\end{equation}
where $Z_{\lambda} = A_c^T W A_c \Delta^{-1}$. Plugging \eqref{X_lambda} into $B_{\lambda}$ gives $B_{\lambda} = [\paren{Z_{\lambda}M_2 M_4^{-1}}^T,\: I_{2 \times 2}]^T$, which implies that $B_{\lambda}^{\perp} = [I_{3 \times 3}, \, -Z_{\lambda}M_2 M_4^{-1}]$. Plugging $X_{\lambda}$ and $B_{\lambda}^{\perp}$ into \eqref{matching2} and using the fact that $\paren{\nabla_q H - \nabla_q \tilde{H}}^T = N - \tilde{N} = \hat{N}$, we have the following solution
\begin{equation}
\label{matching3}
    0 = \begin{bmatrix}
    I - Z_{\lambda} & 0_{3 \times 2}
    \end{bmatrix} [ -\hat{N} + J_2 M^{-1}p + T_{ext}]\ .
\end{equation}

This implies that the first three terms of $-\hat{N} + J_2 M^{-1} p + T_{ext}$, i.e., the unactuated parts, must equal zero to satisfy the matching condition. Then we can obtain the following feasible control law satisfying \eqref{matching3}:
\begin{equation}
\label{controlpolicy}
    u = B^{\dagger}\paren{-\hat{N} + J_2 M^{-1}p + T_{ext}}\ ,
\end{equation}
where $B^{\dagger} = \paren{B^T B}^{-1} B^T$ is the left pseudoinverse of $B$. Moreover, with the unactuated parts of $\hat{N}$ and $Q$ being zero and the actuated parts only depend on actuated state variables, the closed-loop system \eqref{DesHam} is integrable with a well-defined potential function \cite{lin2020optimal}. This ensures the existence of an equivalent Lagragian $\tilde{L} \paren{q, \dot{q}} = \frac{1}{2}\dot{q}^T M \dot{q} + \dot{q}^T Q - \tilde{V}$ that guarantees passivity \cite{blankenstein2002matching}. 

\section{Passivity-Based Data-Driven Controller Design}
\label{sec3}

In this section, we first introduce the controller design that preserves the passivity and stability of the human-exoskeleton system. Then we introduce the data-driven approach that allows the proposed controller resemble normative human joint torque.

\subsection{Passivity and Stability Analysis}

To satisfy the matching condition \eqref{matching3}, the unactuated parts of the controller must equal zero, which restricts the controller to a virtual spring-damper like behavior and prevents it from producing more normative torque. To overcome this limitation, we follow \cite{lin2022optimally} and introduce the new ``power leak" input $T_{ext}$. Define $\hat{N}_{act}$ as the modified gravitational vector that only depends on the actuated parts, then the modified gravitational vector can be expressed as $\hat{N} = \hat{N}_{act} + T_{ext}$, implying that both the actuated components and the external input contribute to $\hat{N}$. Our prior work \cite{lin2022optimally} proved that the human-exoskeleton system is passive with respect to the human input port and the power leak port, and the system is stable in the sense of Lyapunov with certain human inputs. Note that although asymptotic stability can not be guaranteed, Lyapunov stability ensures that the shaped system remains in a neighborhood of the equilibrium point under human's control, which aligns with our objective that the exoskeleton only provides partial assistance and the human is responsible for ensuring stability.

\begin{figure*}[htbp]
\centering
\includegraphics[width = 1\textwidth]{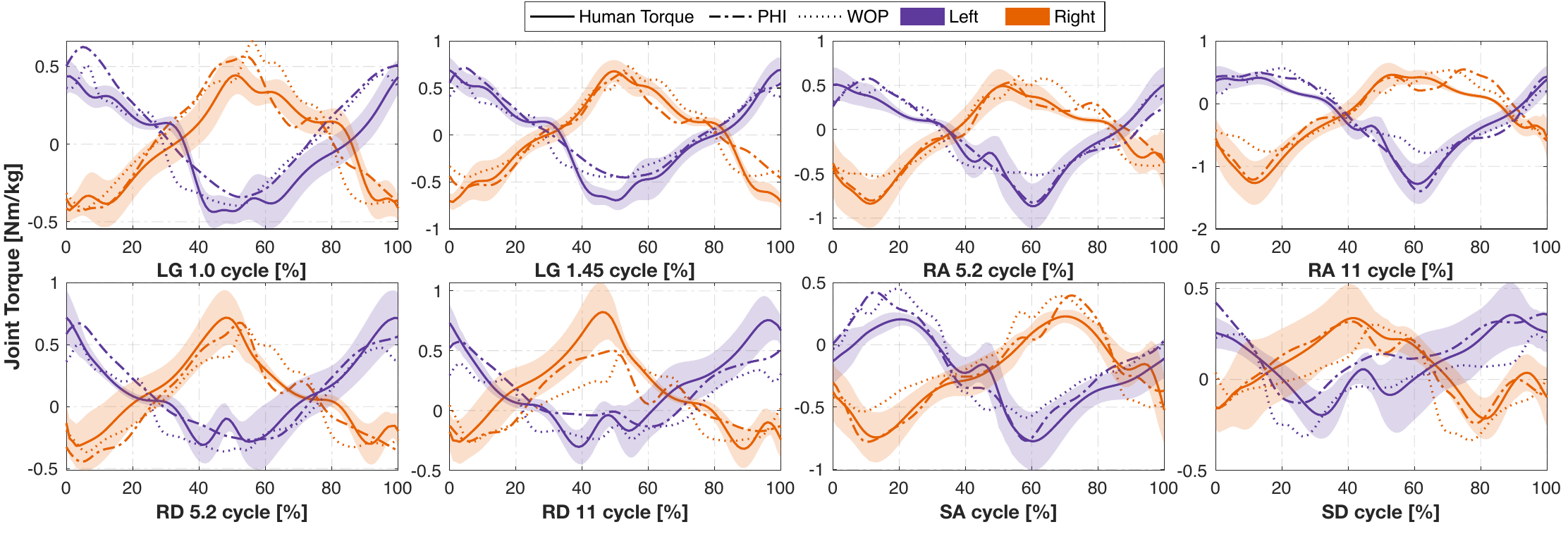}
 \caption{Comparison of across-subject averaged normalized command torque and normalized normative human torque for tasks: level ground walking (LG) at 1.0 and 1.45 m/s, ramp ascending (RA) and ramp descending (RD) with incline $5.2^{\circ}$ and $11^{\circ}$, stair ascending (SA) and descending (SD). The solid lines represent the normative human torque (from \cite{camargo2021comprehensive}), the dotted lines and dash-dotted lines represent command torque generated by WOP mode and PHI mode, respectively. Positive torques represent hip flexion.}
\label{Fig:opt}
\end{figure*}

\subsection{Controller Optimization}

Similar with \cite{lin2022optimally}, we now convert the controller design problem into an optimization problem, where we form the shaping terms in \eqref{controlpolicy} using multiple basis functions. We design $\hat{N} = -\alpha_1 \xi_1 - \cdots - \alpha_i \xi_i$ and $J_2 M^{-1} p = \alpha_{i+1} \xi_{i+1} + \cdots + \alpha_w \xi_w$, where $\Brac{\xi_1, \xi_2, \cdots, \xi_w}$ are the basis functions and $\Brac{\alpha_1, \alpha_2, \cdots, \alpha_w}$ are the coefficients to be solve by optimization. Then the control law in \eqref{controlpolicy} can be rewritten as $u\paren{q, p, \alpha} = B^{\dagger} \paren{\alpha_1 \xi_1 + \cdots + \alpha_w \xi_w}$. To allow the controller produce $u$ that best fits the normative human kinematics in \cite{camargo2021comprehensive}, we design the optimization as
\begin{equation*}
\begin{aligned}
    \arg \min_{\alpha} &\sum_k \sum_j \brac{u\paren{q_{j, k}, p_{j, k}, \alpha} - Y_{j, k}}^T W_{j, k} \big[u\paren{q_{j, k}, p_{j, k}, \alpha} \\
    &- Y_{j, k}\big] + u\paren{0, 0, \alpha}^T W_0\ u\paren{0, 0, \alpha} + \Lambda \| \alpha \|_1 \ ,
\end{aligned}
\end{equation*}
where $u \paren{q_{j, k}, p_{j, k}, \alpha}$ denotes the control law of $k$-th  subject with respect to the $j$-th task, and $Y_{j, k}$ is the corresponding normative data. $W_{j, k}$ and $W_0$ are the weight matrices and $\Lambda$ is the coefficient for $L_1$ regularization. In the objective function, the first term is to minimize the error between control law and the normative data from all subjects for all tasks, while the second term is to minimize the torque provided when $q = p = 0$, i.e., the user is standing straight. The $L_1$ regularization term is included to promote sparsity and prevent overfitting. CVX, a package designed for convex optimization, was used to solve the problem \cite{cvx}.

In this paper, two shaping strategies are considered: 1) Hamiltonian without $\phi$ (WOP), where basis functions only depends on $\theta_l$ and $\theta_r$, and 2) Hamiltonian with $\phi$ (PHI), where $\phi$ is also considered to form the basis functions. We fit the controller output to the normative human torque of 10 subjects from tasks including level ground walking at 1.0 (LG 1.0) and 1.4 m/s (LG 1.4), ramp ascent/descent with $5.2^{\circ}$ (RA/RD 5.2), $11^{\circ}$ (RA/RD 11) inclines and stairs ascent/descent with 4 inches step height (SA/SD) in \cite{camargo2021comprehensive} to obtain the optimal coefficients $\alpha^*$, which will also be used in the human subject experiments. The optimization result is shown in Fig. \ref{Fig:opt}. Note that the positive normative torque for stair ascending is scaled up to provide more flexion assistance. Overall, PHI fits normative data better due to the extra flexibility provided by global thigh angle $\phi$ in basis functions. For certain tasks like ramp ascending and stair ascending, PHI provides more extension assistance, while for stair descending PHI provides more flexion assistance.

To compare the two methods used in our optimization, we use two metrics: Cosine Similarity (SIM) and Variance Accounted For (VAF) as defined in \cite{lin2022optimally}:
 \begin{align*}
     & \mathrm{SIM}(A, B) = \frac{100 \cdot A \cdot B}{\|A\|_2 \|B\|_2}\ , \\
     & \mathrm{VAF}(A, B) = 100 \cdot \brac{1 - \frac{\mathrm{var}(A-B)}{\mathrm{var}(A)}}\ .
 \end{align*}
The right and left sides are averaged together. Leave-one-subject-out cross-validation with ten subjects in total was performed to check the performance of proposed method in the presence of inter-subject differences. As shown in TABLE \ref{tab1}, both methods perform well for all the tasks with PHI being sightly better than WOP.

\begin{table}[htbp]
\caption{SIM and VAF comparison: Showing Mean $\paren{\pm \mathrm{SD}}$ for different tasks (rows) and methods (columns)}
\begin{center}
\begin{tabular}{|c|c|c|c|c|}
\hline
\textbf{Task} & \multicolumn{2}{|c|}{\textbf{SIM [\%]}} & \multicolumn{2}{|c|}{\textbf{VAF [\%]}} \\
\cline{2-5}
\textbf{Names} & \textbf{PHI} & \textbf{WOP} & \textbf{PHI} & \textbf{WOP} \\
\hline
LG 1.0& \textbf{86.5 (7.9)}& 82.7 (8.0) & \textbf{71.4 (17.7)}& 62.9 (22.1) \\
\hline
LG 1.45& \textbf{91.5 (2.5)}& 89.9 (3.3)& \textbf{82.5 (4.8)}& 79.0 (6.1) \\
\hline
RA 5.2& \textbf{87.3 (8.5)}& 83.6 (9.3)& \textbf{71.2 (12.1)}& 63.5 (14.9) \\
\hline
RA 11& \textbf{89.7 (3.9)}& 86.2 (5.6)& \textbf{73.8 (9.1)}& 71.0 (13.3) \\
\hline
RD 5.2& \textbf{84.7 (5.7)}& 81.3 (6.5)& \textbf{70.7 (12.8)}& 64.9 (16.1) \\
\hline
RD 11& \textbf{74.6 (11.6)}& 72.0 (12.7)& 52.3 (13.5)& \textbf{57.8 (9.8)} \\
\hline
SA& \textbf{86.2 (11.4)}& 80.4 (15.2)& \textbf{62.5 (20.7)}& 47.2 (35.5) \\
\hline
SD& \textbf{65.6 (16.8)}& 59.4 (11.0)& \textbf{55.6 (22.6)}& 45.6 (21.3) \\
\hline
\end{tabular}
\label{tab1}
\end{center}
\end{table}

\section{Proof of Concept Experiment with Able-Bodied Human Subjects}
\label{sec4}

In this section, we introduce the implementation of the proposed controller on a commercial hip exoskeleton. Then we present the able-bodied human subject experiment, which acts as a proof-of-concept to demonstrate the controller's ability to assist multiple tasks.

\subsection{Hardware Implementation}

The proposed controller was implemented on the \textit{Movex} hip exoskeleton (Enhanced Robotics, Shenzhen, China) shown in Fig. \ref{Fig:model}. 
\textit{Movex} has two hip joint actuators which provide 12.8 Nm continuous torque using 125 W brushless DC motors (ER-6510-S, Enhanced Robotics) with a 32:1 transmission ratio. This relatively low transmission ratio ensures backdrivability. The joint angles are measured using magnetic incremental encoders. Since \textit{Movex} does not have an inertial measurement unit (IMU), a 9-axis IMU (3DM-CX5-25, LORD MicroStrain) was added to measure the global thigh angle. The low-level motion control runs on an custom designed STM32F405RG microcontroller (168 MHz, STMicroelectronics) and the high-level controller is implemented on a Raspberry Pi 3B+ (1.4 GHz 64-bit processor, 1 GB RAM, Raspberry Pi Foundation). The whole system is powered by a 25.2V, 66.7 Wh Li-ion battery (LiPo 18650 6S1P, Samsung), which allows at least 4 hr of operation. The weight of the whole system is about 4 kg. The assistance torques provided are determined by multiplying the optimized command torque (in Nm/kg) with the subjects' body mass and the level-of-assistance (\% LOA). Software and hardware saturation for output torque are set to ensure the safety of human subjects and motors.

\subsection{Experiment Method}

As a proof-of-concept, two healthy human subjects were enrolled (s1: male, mass: 80 kg, height: 1.78 m; s2: male, mass: 62 kg, height: 1.67 m) to 1) demonstrate the capability of proposed controller to assist ADL and 2) compare the performance of WOP and PHI. The study was approved by the Institutional Review Board at the University of Michigan (HUM00164931). The muscle activation of Rectus Femoris (RF), Biceps Femoris (BF) and Gluteus Maximus (GM) was accessed via EMG (Trigno Avanti Sensor, Deslys Inc.), where RF functions as a hip flexor and BF, GM function as hip extensors \cite{neumann2010kinesiology}. 

The experiment comprised five tasks: level ground walking, $\pm 6$ deg ramp ascent/descent on an instrumented treadmill (Bertec Corporation), and stair ascent/descent over 6-inch steps. The speed of walking were determined by the subjects and a metronome was set with self-selected beats-per-minute (BPM) to help subjects maintain their cadence. Before the experiments, subjects were provided with training sessions to acclimate to the exoskeleton and decide the speed, BPM and $\%$LOA. Subjects were allowed to use handrails to prevent falling in the training session, but in experiments using handrails was disallowed. For s1, the speed and BPM for each task were: level ground (0.9 m/s with 95 BPM), ramp ascent (0.65 m/s with 75 BPM), ramp descent (0.65 m/s with 90 BPM), and stairs (80 BPM). For s2, the values were: level ground (0.9 m/s with 105 BPM), ramp (0.8 m/s with 100 BPM), and stairs (95 BPM).  The \%LOA for s1 and s2 was $40\%$ and $35\%$, respectively. The video of experiment is provided as supplemental material at \cite{video}.

All the tasks were repeated for three modes: bare (without exoskeleton), active exoskeleton with $\phi$ (PHI), and active exoskeleton without $\phi$ (WOP). For the treadmill trials, the subject walked for 3 min, and data was collected during the last 2 min for relatively consistent kinematics. For the trials on stairs, subjects first ascended a staircase with 19 steps, stopped, turned around and then descended the staircase, which was repeated three times for each trial. Breaks lasting at least 2 min were provided between trials to prevent fatigue and were extended at the request of the subject.

All trials were cropped into gait cycles by heel strikes detected by a heel-mounted IMU (not used for feedback control). The EMG for each muscle was first bandpass filtered by a fourth-order Butterworth filter (20-200 Hz) and smoothed with a 100 ms moving window RMS, then normalized with respect to the maximum peak of the ensemble averages of the three exoskeleton modes (separately for each task and muscle). This converted the signals to a percentage of the maximum voluntary contraction level (\% MVC), which was then integrated along time to obtain muscular effort represented as \% MVC.s, similar to \cite{divekar2020potential}. According to the result of Shapiro-Wilk normality test, the EMG data were not normally distributed, so we performed the Kruskal–Wallis test with a Bonferroni correction to check if the difference of muscle efforts was statistically significant. When $p < 0.05$, the null hypothesis was rejected.

\subsection{Experiment Result}

\begin{figure}[htbp]
\centering
\includegraphics[width = 0.47\textwidth]{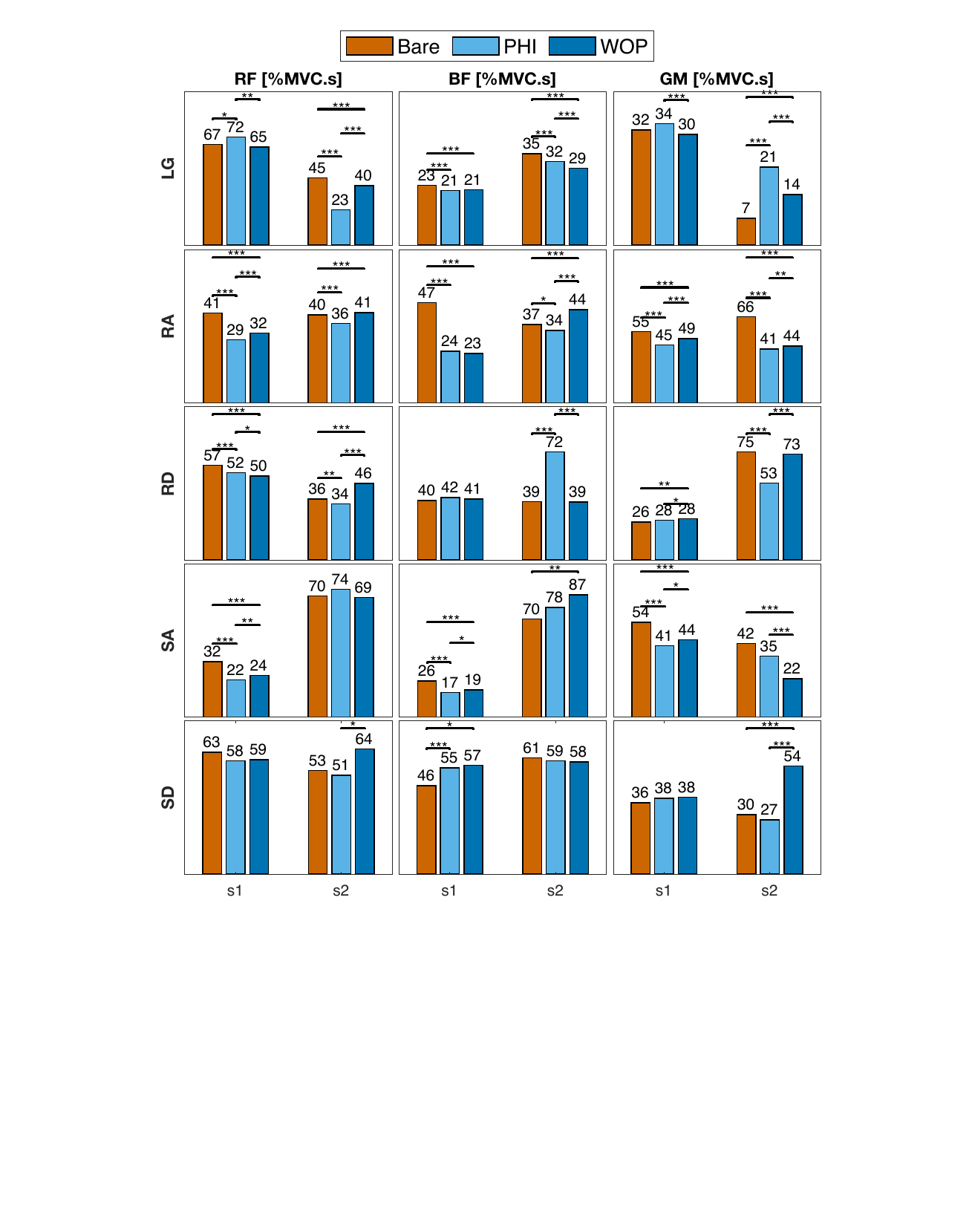}\
\caption{Individual subject comparison of mean effort across repetitions for each muscle pair (RF, BF and GM) with bare, WOP and PHI modes. * represents statistical significance with $p < 0.05$, ** represents $p < 0.01$ and *** represents $p < 0.001$. }
\label{Fig:stat}
\end{figure}

Fig. \ref{Fig:stat} presents the muscular effort comparison between different modes for the five tasks. For most tasks the active modes (PHI and WOP) generally reduced RF effort, which is the dominant muscle of swing phase. For s1, the strongest effort reduction appeared in RA and SA, which both require much hip flexion, while for s2 this appeared in RA and LG. For the stance phase, the dominant muscles are BF and GM, which also exhibited effort reduction in the active mode. For BF, effort reduction appeared in LG and RA for both s1 and s2, while for s1 the BF effort also reduced in SA. For GM, effort reduction appeared in RA and SA for both subjects, and for s2 the GM effort also reduced in RD. However, for SD the effect on muscle effort by the active exoskeleton was not statistically significant. 

\begin{figure}[htbp]
\centering
\includegraphics[width = 0.47\textwidth]{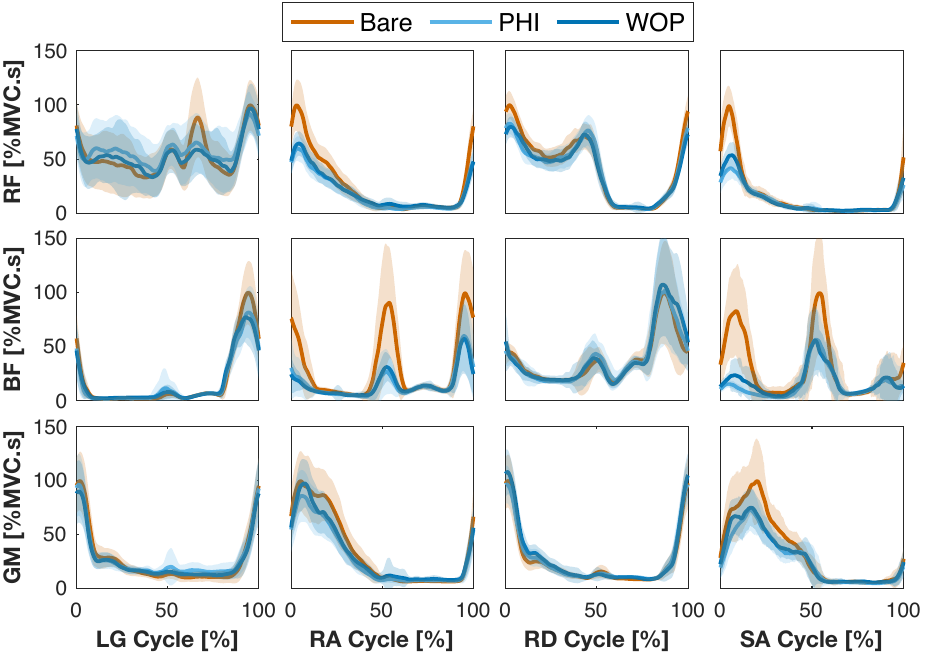}\
\caption{Subject 1 EMG comparison between bare, PHI, WOP modes for each muscle (RF, BF and GM) for four tasks (LG, RA, RD, SA). Solid lines are the time-normalized ensemble averages across gait cycles.}
\label{Fig:effort}
\end{figure}

There were also differences between WOP and PHI. In RA (s1 and s2) and SA (s1), the hip extensors (BF and GM) had less effort in the PHI mode than in the WOP mode, which also appeared in SD for the hip flexior (RF). This aligns with the optimization result in Fig. \ref{Fig:opt}, which implies that PHI can provide more assistance for certain tasks due to the extra flexibility provided by the global thigh angle $\phi$. Fig. \ref{Fig:effort} presents the ensemble-averaged EMG of RF, BF and GM of s1 for bare and active modes with respective to LG, RA, RD and SA tasks, where we also see significant reduction in muscle effort and peak EMG compared with the bare mode.


Overall, the observations in Fig. \ref{Fig:stat} and Fig. \ref{Fig:effort} align with the optimization result in Fig. \ref{Fig:opt} and demonstrate the potential for assisting multiple tasks. However, for some tasks, the active mode caused more muscle effort, suggesting that more resistance than assistance was provided. For example, the muscle effort of s1 slightly increased for RF and GM in LG and for GM in RD, which is possibly caused by the backdrive torque (about 0.7 Nm) and extra weight added to the lower-limb. While this can be partially solved by compensating backdrive torque as in \cite{weiss2012feed}, the active mode muscle effort of s2 significantly increased for GM in LG and BF in RD, which suggests the control torque resisted the volitional motion of subject. This inter-subject difference indicates that additional human subjects would be needed to make more general conclusions about the effectiveness of the controller, which is left to our future work.


\section{Conclusion}
\label{sec5}
In this paper, we proposed an energy shaping-based control strategy for a hip exoskeleton. Based on the IDA-PBC method, the proposed controller provides both BWS and velocity-dependent assistance. Analysis showed that the system is stable and preserves the passivity of the human-exoskeleton system. Optimization was used to fit the controller output into normative human torques. A proof-of-concept experiment with two able-bodied human subjects was performed to demonstrate the proposed controller's ability of assist ADL. More extensive clinical testing with the proposed controller will be included in our future work, and this control method will also be extended to different hip exoskeletons like M-BLUE \cite{nesler2022enhancing}.

\section*{Acknowledgement}
The authors gratefully thank Nikhil V. Divekar, Avani Yerva and Land Burnam for their assistance.




\bibliographystyle{IEEEtran}
\bibliography{IEEEexample}

\end{document}